Krzysztof WOŁK
Emilia REJMUND
Krzysztof MARASEK
Polsko-Japońska Akademia Technik Komputerowych, Warszawa

# Multi-domain machine translation enhancements by parallel data extraction from comparable corpora

Poprawa jakości tłumaczenia maszynowego dla wielu domen
poprzez ekstrakcję danych równoległych z korpusów porównywalnych

**Streszczenie**

Teksty równoległe są zasobem językowym spotykanym stosunkowo rzadko, jednak stanowią bardzo użyteczny materiał badawczy o szerokim zastosowaniu np. podczas międzyjęzykowego wyszukiwania informacji oraz w statystycznym tłumaczeniu maszynowym. Niniejsze badanie prezentuje i analizuje opracowane przez nas nowe metody pozyskiwania danych z korpusów porównywalnych. Metody te są automatyczne i działają w sposób nienadzorowany, co czyni je użytecznymi w budowie korpusów równoległych na szeroką skalę. W niniejszym badaniu proponujemy metodę automatycznego przeszukiwania sieci w celu zbudowania korpusów porównywalnych zrównoleglonych na poziomie tematu, np. na podstawie danych z Wikipedii czy strony Euronews.com. Opracowaliśmy również nowe metody pozyskiwania równoległych zdań z danych porównywalnych oraz proponujemy metody filtracji korpusów równoległych zdolne selekcjonować niezgodne ze sobą lub tylko częściowo ekwiwalentne pary zdań. Za pomocą naszych metod można pozyskać zasoby równoległe dla dowolnej pary języków. Ewaluację jakości zbudowanych korpusów przeprowadzono poprzez analizę wpływu ich użycia na systemy statystycznego tłumaczeniu maszynowego przy wykorzystaniu typowych miar jakości tłumaczenia. Eksperymenty zostały zaprezentowane na przykładzie pary językowej polski-angielski dla różnego typu tekstów, tj. wykładów, rozmówek turystycznych, dialogów filmowych, zapisów posiedzeń Europarlamentu oraz tekstów zawartych w ulotkach leków. Przetestowaliśmy także drugą metodę tworzenia korpusów równoległych na podstawie danych z korpusów porównywalnych, pozwalającą automatycznie poszerzyć istniejący korpus zdań z danej tematyki, wykorzystując znalezione między nimi analogie. Metoda ta nie wymaga posiadania wcześniejszych zasobów równoległych celem stworzenia i dostosowania klasyfikatora.

Wyniki naszych eksperymentów są obiecujące. Z artykułów Wikipedii udało się pozyskać prawie pół miliona zdań równoległych i niespełna 5.000 z portalu Euronews.com (z wykorzystaniem pierwszej z metod) oraz 114.000 z Wikipedii, wykorzystując analogie między artykułami. Pozyskane dane wpłynęły pozytywnie na jakość tłumaczenia maszynowego, która została zmierzona popularnymi miarami automatycznymi tj. BLEU, NIST, TER oraz METEOR. Jednak dane pozyskane automatycznie po manualnej analizie okazały się „zaszumione", dlatego też podjęto próbę ich automatycznego przefiltrowania. Metodę filtrowania danych zbadano, porównując jej wyniki z wynikami uzyskanymi przy zastosowaniu metody polegającej na ocenie ludzkiej, a także badając jej wpływ na tłumaczenie maszynowe. Filtrowanie okazało się skuteczne, gdyż polepszyło ostateczne wyniki statystycznego tłumaczenia maszynowego.

**1 Introduction**

Parallel sentences are an invaluable information resource especially for machine translation systems as well as for other cross-lingual information-dependent tasks. Unfortunately, such a type of data is quite rare, even for the Polish-English language pair. On the other hand, monolingual data for those languages is accessible in far greater quantities. We can classify the similarity of data as four main corpora types (Wu, Fung, 2005). The most rare is a parallel corpus. It is a collection of texts, each of which is translated into one or more languages other than the original. Such data should be aligned at least at the sentence level. A noisy-parallel corpus contains bilingual sentences that are not perfectly aligned or which have not been precisely translated. Nevertheless, they should mostly contain translations of specific phrases within a document. A comparable corpus is built from non-sentence-aligned and not-translated bilingual documents, but the documents should be topic-aligned. A quasi-comparable corpus includes very heterogeneous and very non-parallel bilingual documents that can – but do not have to – be topic-aligned (Wu, Fung, 2005).

In this article we present methodologies that allow us to obtain truly parallel corpora from data sources, which have not been sentence-aligned, such as noisy-parallel or comparable corpora. For this purpose, we used a set of specialized tools for obtaining, aligning, extracting and filtering text data, combined together into a pipeline that allows us to complete the task. We present the results of our initial experiments based on text samples obtained from Wikipedia dumps and the Euronews web page. We chose Wikipedia as a source of data because of a large number of documents that it provides (1,047,423 articles on PL Wiki and 4,524,017 on EN Wiki at the time of writing this article). Furthermore, Wikipedia contains not only comparable documents, but also some documents that are translations of each other. The quality of our approach is measured by improvements in machine translation (MT) results.

The second method is based on sequential analogy detection. We seek to obtain parallel corpora from unaligned data. Such an approach was presented in literature (Koehn, Haddow, 2012; Chu, Nakazawa, Kurohashi, 2013), but all applications concern similar languages with similar grammars like English-French, Chinese-Japanese. We try to apply this method for English-Polish corpora. These two languages have different grammar, which makes our approach innovative and can easily be adapted for different languages pairs. In our approach, to enhance the quality of identified analogies, sequential analogy clusters are sought.

## 2 State of the Art

The development on Statistical Machine Translation (SMT) systems for Polish has progressed slower than for other more popular languages in recent years. The tools used for mainstream languages were not adapted for Polish. As far as comparable corpora are concerned, many attempts have been made (especially for Wikipedia), but none of them for the Polish language.

Two main approaches for building comparable corpora can be distinguished. Probably the most common approach is based on the retrieval of cross-lingual information from texts. In the second approach, source documents need to be translated using any machine translation system. The documents translated in that process are then compared with documents written in the target language in order to find the most similar document pairs.

Skadiņa and Aker (2006) suggested obtaining only the title and some meta-information, such as publication date and time for each document instead of its full contents in order to reduce the cost of building the comparable corpora (CC). The cosine similarity of title term frequency vectors were used to match titles and contents of matched pairs.

An interesting idea for mining parallel data from Wikipedia was described in Adafree and de Rijke (2014). The authors propose two separate approaches. The first idea is to use an online machine translation (MT) system to translate Dutch Wikipedia pages into English and they try to compare original EN pages with the translated ones. The idea, however interesting, is most likely computationally unreasonable and this is an example of the chicken-and-egg problem.

The second idea uses a dictionary generated from Wikipedia titles and shared hyperlinks between documents. Unfortunately, the second method was reported to return numerous noisy sentence pairs.

Kilgarriff, Avinesh and Pomikalek (2011) improve the BootCat method that was proven to be fast and effective as far as corpus building is concerned. The authors try to extend this method by adding support for multilingual data and also present a pivot evaluation.

Interwiki links were utilized by Tyer and Pienaar (2008). Based on the Wikipedia link structure a bilingual dictionary is extracted. In their work they measured the mismatch between linked Wikipedia pages. They found that their precision is about 69-92% depending on a language.

Smith, Quirk and Toutanova (2010) try to advance the state of the art in parallel data mining by modeling document level alignment using the observation that parallel sentences can most likely be found in close proximity. They also use annotation available on Wikipedia and an automatically induced lexicon model. The authors report precision of about 90 percent.

What is more Pal, Pakray and Naskar (2014) introduce an automatic alignment method of parallel text fragments by using a textual entailment technique and a phrase-base Statistical Machine Translation (SMT) system. The authors state that a significant improvement in SMT quality was obtained (an increase in BLEU by 1.73) by using mined data.

Strotgen and Gertz (2012) introduce a document similarity measure that is based on events. In order to count the values of this metric, documents are modeled as sets of events that are temporal and geographical expressions are found in the documents. Target documents are ranked based on temporal and geographical hierarchies.

In this research a Yalign tool is used (described in detail in section 4.1). The solution is far from perfect but after improvements that were made during this study, it supplied the SMT systems with bi-sentences of good quality in a reasonable amount of time.

## 3 Preparation of the data

Our procedure starts with a specialized web crawler implemented by us. Because PL Wiki contains less data of which almost all articles have their corresponding entries on EN Wiki, the program crawls data starting from the non-English site first. The crawler can obtain and save bilingual articles of any language supported by Wikipedia. The tool requires at least two Wikipedia dumps in different languages and information about language links between the articles in the dumps. For Euronews.com another web crawler was used. It generates a database of parallel articles in two selected languages in order to collect comparable data from it.

Before a mining tool processes the data the texts must be prepared. First, all the data is saved in a database. Secondly, the tool aligns pairs of articles and removes the articles that do not exist in both languages from the database. Such topic-aligned articles are filtered in order to remove any HTML, XML tags or noisy data (tables, references, figures, etc.). Finally, bilingual documents are tagged with a unique ID and form a topic-aligned comparable corpus.

For the experiments in statistical machine translation we choose the domain of TED lectures, specifically the PL-EN TED[1] corpora prepared for the IWSLT (International Workshop on Spoken Language Translation) 2013 evaluation campaign by the FBK (Fondazione Bruno Kessler). This domain is very wide and covers many subjects and areas. The data contains almost 2,5M untokenized words (Cetollo, Girardi, Federico, 2012). Additionally, we choose two more narrow domains: The first parallel corpus is made out of PDF documents from the European Medicines Agency (EMEA) and medicine leaflets (Tiedemann, 2009). The second was extracted from the proceedings of the European Parliament (EUP) (Tiedemann, 2012). We also conducted experiments on the Basic Travel Expression Corpus (BTEC), a multilingual speech corpus containing tourism-related sentences similar to those that are usually found in phrasebooks for tourists going abroad (Marasek, 2012). Lastly, we used a corpus built from the movie subtitles (OPEN) (Tiedemann, 2009). Table 1 presents details of the numbers of unique words (WORDS) and their forms as well as of the numbers of bilingual sentence pairs (PAIRS) in each of the corpora.

| CORPORA | PL WORDS | EN WORDS | PAIRS |
|---|---|---|---|
| BTEC | 50,782 | 24,662 | 220,730 |
| TED | 218,426 | 104,117 | 151,288 |
| EMEA | 148,230 | 109,361 | 1,046,764 |
| EUP | 311,654 | 136,597 | 632,565 |
| OPEN | 1,236,088 | 749,300 | 33,570,553 |

Table 1. Corpora specification

As mentioned, our procedure can be divided into three main steps. First the data is collected, then it is aligned at the article level, and lastly the results of the alignment are mined for parallel sentences. The last two steps are not trivial because of the disparities between Wikipedia documents. Based on the Wikipedia statistics we know that an average article on PL

---
[1] https://www.ted.com/talks

Wiki contains about 379 words, whereas on EN Wiki it has 590 words. The corpus might also contain imprecise or indirect translations or totally new texts making the alignment difficult. Thus, alignment is crucial for accuracy of mining process. Sentence alignment must also be computationally feasible in order to be of practical use in various applications.

The Polish language presents a particular challenge to the application of such tools. It is a complicated West-Slavic language with relatively complex lexical elements and complicated grammatical rules. In addition, Polish has a large vocabulary due to prefixes and many endings representing word declension. These characteristics have a significant impact on the data and data structure requirements.

In contrast, English is a position-sensitive language. The syntactic order (the order of words in a sentence) plays a significant role, and inflection of words is limited (due to the lack of declension endings). The position of a word in an English sentence is often the only indicator of its function. The sentence order follows the Subject-Verb-Object (SVO) schema, with the subject phrase preceding the predicate. On the other hand, no specific word order is imposed in Polish, and the word order has little effect on the meaning of a sentence. The same idea can be expressed in several ways. It must be noted that such differences exist in many language pairs and need to be dealt with in some way (Wołk. Marasek, 2013a).

With this methodology we were able to obtain 4,498 topic-aligned articles from Euronews and 492,906 from Wikipedia.

**4 Parallel data mining**

In order to extract parallel sentence pairs, we decided to try two different strategies. The first one is facilitated by the Yalign tool[2] and the second is based on analogy detection. The MT results we present in this article were obtained with the first strategy. The second method is still in its development phase, nevertheless the initial results are promising and worth mentioning.

**4.1 The Yalign tool**

The Yalign tool was designed in order to automate the parallel text mining process by finding sentences that are close translation matches from the comparable corpora. This opens up avenues for harvesting parallel corpora from comparable sources like bilingual documents and the web. What is more, Yalign is not limited to any language pair, however the creation of unique alignment models for two required languages is necessary.

---
[2] https://github.com/machinalis/yalign

The Yalign tool was implemented using a sentence similarity metric that produces a rough estimate (a number between 0 and 1) of how likely it is for two sentences to be a translation of each other. Additionally, it uses a sequence aligner that produces an alignment that maximizes the sum of the individual (per sentence pair) similarities between two documents. Yalign's algorithm is actually a wrapper before the standard sequence alignment algorithm[3].

For the sequence alignment, Yalign uses a variation of the Needleman-Wunch algorithm[4] (originally used for DNA sequences) to find an optimal alignment between the sentences in two given documents. The algorithm has polynomial time worst-case complexity and it produces an optimal alignment. Unfortunately, it cannot handle alignments that cross each other or alignments from two sentences into a single one[4].

Since the sentence similarity calculation is a computationally-expensive operation, the implemented variation of the Needleman-Wunch algorithm uses the A* approach to explore the search space instead of using the classical dynamic programming method that would require N * M calls to the sentence similarity matrix.

After the alignment, only sentences that have a high probability of being translations of each other are included in the final alignment. The result is filtered in order to deliver high quality alignments. To do this, a threshold value is used, such that if the sentence similarity metric is too low, the pair is excluded. For the sentence similarity metric, the algorithm uses a statistical classifier's likelihood output and adapts it into the <0,1> range. The classifier must be trained in order to determine if a pair of sentences is a translation of each other or not. The particular classifier used in the Yalign project was the Support Vector Machine (SVM). Besides being an excellent classifier, SVMs can provide a distance to the separation hyperplane during classification, and this distance can be easily modified using the Sigmoid Function to return the likelihood between 0 and 1 (Thorsten, 2005). The use of a classifier means that the quality of the alignment depends not only on the input but also on the quality of the trained classifier.

Unfortunately, the Yalign tool is not computationally feasible when large-scale parallel data mining is concerned. The standard implementation accepts as input plain text or web links that need to be accepted, and for each pair alignment the classifier is loaded into memory. In addition, Yalign is single-threaded. In order to improve the performance, we developed a solution that supplies Yalign tool with articles from the database within one session, with no need to reload the classifier each time. What is more, our solution facilitated multithreading and proved to increase the mining time by the factor of 5 (using a 4 core, 8 thread Core i7 CPU).

---

[3] http://yalign.readthedocs.org/en/latest/
[4] https://www.cs.utoronto.ca/~brudno/bcb410/lec2notes.pdf

To train the classifier, a good quality parallel data was necessary as well as a dictionary with translation probabilities included. For this purpose, we used TED talks (Cetollo, Girardi, Federico, 2012) corpora enhanced by us during the IWSLT'13 Evaluation Campaign (Wołk, Marasek, 2013a). In order to obtain a dictionary, we built a phrase table and extracted 1-grams from it. We used the MGIZA++ tool for word and phrase alignment. The lexical reordering was set to use the msd-bidirectional-fe method and the symmetrisation method was set to grow-diag-final and for word alignment processing (Wołk, Marasek, 2013). We used the four previously-described corpora as bilingual training data. We obtained four different classifiers and repeated mining procedure with each of them.

Using this method, we successfully mined about 80MB corpora from Wikipedia and 0,3MB from Euronews. Each of the parallel data sets were combined together into one big corpus on which the MT experiments were conducted. The detailed results for Wikipedia are presented in Table 2.

| Classifier | Value | PL | EN |
|---|---|---|---|
| TED | Size in MB | 41,0 | 41,2 |
| | No. of sentences | 357,931 | 357,931 |
| | No. of words | 5,677,504 | 6,372,017 |
| | No. of unique words | 812,370 | 741,463 |
| BTEC | Size in MB | 3,2 | 3,2 |
| | No. of sentences | 41,737 | 41,737 |
| | No. of words | 439,550 | 473,084 |
| | No. of unique words | 139,454 | 127,820 |
| EMEA | Size in MB | 0,15 | 0,14 |
| | No. of sentences | 1,507 | 1,507 |
| | No. of words | 18,301 | 21,616 |
| | No. of unique words | 7,162 | 5,352 |
| EUP | Size in MB | 8,0 | 8,1 |
| | No. of sentences | 74,295 | 74,295 |
| | No. of words | 1,118,167 | 1,203,307 |
| | No. of unique words | 257,338 | 242,899 |
| OPEN | Size in MB | 5,8 | 5,7 |
| | No. of sentences | 25,704 | 25,704 |

|  |  |  |
|---|---|---|
| No. of words | 779,420 | 854,106 |
| No. of unique words | 219,965 | 198,599 |

Table 2. Data mined from Wikipedia for each classifier

During the empirical research we realized that, as in the case of machine translation in which different results and quality measures are obtained depending on whether the system was trained from foreign to native language or opposite, Yalign suffers from a similar problem. In order to cover as much parallel data as possible during mining, it is also necessary to train the classifiers bidirectionally as far as specific language pairs are concerned. By doing so, additional bi-sentences can be found. Some of them will be repeated, however, in our opinion, the potential increase of the size of parallel corpora is worth that effort. Table 3 demonstrates how many sentences were obtained in the second phase of mining as well as how many of them were overlapping. The number of additionally-mined data is counted as well.

| **Classifier** | **Value** | **Data Mined** |
|---|---|---|
| TED | Recognized sentences | 132,611 |
|  | Overlapping sentences | 61,276 |
|  | Newly obtained | 71,335 |
| BTEC | Recognized sentences | 12,447 |
|  | Overlapping sentences | 9,334 |
|  | Newly obtained | 3,113 |
| EMEA | Recognized sentences | 762 |
|  | Overlapping sentences | 683 |
|  | Newly obtained | 79 |
| EUP | Recognized sentences | 23, 952 |
|  | Overlapping sentences | 21,304 |
|  | Newly obtained | 2,648 |
| OPEN | Recognized sentences | 11,751 |
|  | Overlapping sentences | 7,936 |
|  | Newly obtained | 3,815 |

Table 3. Corpora statistics obtained in the second mining phase

## 4.2. The analogy-based method

This method is based on sequential analogy detection. Based on a parallel corpus we detect analogies that exists between both languages. In order to enhance the quality of identified analogies, sequential analogy clusters are sought.

However, our current research on the Wikipedia corpora shows that it is both extremely difficult and machine-time-consuming to seek clusters of higher orders. Therefore, we limited our search to simple analogies such as A is to B in the same way as C is to D.

**A:B::C:D**

Such analogies are found using distance calculation. We seek such sentences that:

**dist(A,B)=dist(C,D)**

and

**dist(A,C)=dist(B,D)**

An additional constrain was added that requires the same relation of occurrences of each character in the sentences. For example, if the number of character "a" in sentence A is equal to x and equal to y in sentence B then the same relation must occur in sentences C and D.

We used the Levenshtein metric in our distance calculation. We tried to apply it directly to the characters in a sentence, or consider each word in a sentence as an individual symbol, and calculate the Levenshtein distance between symbol-coded sentences. The latter method was employed because it had earlier been tested on the Chinese and Japanese languages (Yang, Lepage, 2014) which use symbols to represent entire words.

After clustering, the data from clusters are compared to each other to find similarities between them. For each four sentences

**A:B::C:D**

we look for such E and F that:

**C:D::E:F and E:F::A:B**

However, no such sentences were found in our corpus, therefore we limited our analysis to small clusters of the size of 2 pairs of sentences. In every cluster, matching sentences from the parallel corpus were identified. It let us generate new sentences similar to the ones which are in our corpus and add them to the resulting data set. For each of sequential analogies which were identified, a rewriting model is constructed. This is achieved by string manipulation. Common prefixes and suffixes for each of the sentence pairs are calculated using the LCS (Longest Common Subsequence) method.

A sample of the rewriting model is shown in this example (the prefix and the suffix are shown in bold)

***Poproszę** koc i poduszkę.* ⇔ *A blanket and a pillow**, please.***

***Czy mogę poprosić** o śmietankę i cukier?* ⇔ ***Can I have** cream and sugar?*

The rewriting model consists of a prefix, a suffix and their translation. It is now possible to construct a parallel corpus form a non-parallel monolingual source. Each sentence in the corpus is tested for a match with the model. If a sentence contains a prefix and a suffix, it is considered a matching sentence.

***Poproszę** bilet.* ⇔ *A unknown**, please.***

In the matched sentence some of the words remain untranslated but the general meaning of the sentence is conveyed. Remaining words may be translated word-by-word while the translated sentence will remain grammatically correct.

bilet ⇔ ticket

By substituting unknown words with their translations, we are able to create a parallel corpus entry.

***Poproszę** bilet.* ⇔ *A ticket**, please.***

As a result of the sequential analogy-detection-based method we mined 8,128 models from our Wikipedia parallel corpus. This enabled us to generate 114,000 new sentence pairs to build a parallel corpus. The sentences were generated from the Wikipedia comparable corpus that contains extracts of Wikipedia articles. Therefore, we have articles in Polish and English on the same topic, but sentences are not aligned in any particular way. We use rewriting models to match sentences from the Polish article to sentences in English. Whenever the model can be successfully applied to a pair of sentences, this pair is considered to be parallel resulting in the generation of a quasi-parallel corpus ('quasi', since the sentences are aligned artificially using the approach described above). These parallel sentences can be used to extend parallel corpora in order to improve the quality of the SMT system.

**5 Evaluation**

In order to evaluate the corpora, we divided each corpus into 200 segments and randomly selected 10 sentences from each segment. This methodology ensured that the test sets covered the entire corpus. The selected sentences were removed from the corpora. We trained the baseline system, as well as the system with extended training data with the Wikipedia corpora and next we used Modified Moore Levis Filtering for the Wikipedia corpora domain adaptation. Additionally, we used the monolingual part of the corpora as a language model and we tried to adapt it for each corpus by using linear interpolation (Koehn, Haddow, 2012).

Summing up, the evaluation was done using test sets built from 2,000 randomly selected bi-sentences taken from each domain. For scoring purposes we used four well-known metrics that show high correlations with human judgments. Among the commonly used SMT metrics are: Bilingual Evaluation Understudy (BLEU), the U.S. National Institute of Standards & Technology (NIST) metric, the Metric for Evaluation of Translation with Explicit Ordering (METEOR), and Translation Error Rate (TER).

According to Tiedemann (2012) BLEU uses textual phrases of varying length to match SMT and reference translations. The scoring with this metric is determined by the weighted averages of these matches.

To evaluate infrequently-used words, the NIST (Wołk, Marasek, 2014a) metric scores the translation of such words higher and uses the arithmetic mean of n-gram matches. Smaller differences in phrase length incur a smaller brevity penalty. This metric has shown advantages over the BLEU metric.

The METEOR (Wołk, Marasek, 2014a) metric also changes the brevity penalty used by BLEU, uses the arithmetic mean like NIST, and considers matches in word order through examination of higher order n-grams. These changes increase the score based on recall. This metric also considers best matches against multiple reference translations when evaluating the SMT output.

TER (Wołk, Marasek, 2014a) compares the SMT and reference translations to determine the minimum number of edits a human would need to make the sentence pairs equivalent in both fluency and semantics. The closest match to a reference translation is used in this metric. There are several types of edits considered: word deletion, word insertion, word order, word substitution, and phrase order.

## 6. Experimental results

A set of experiments was performed to evaluate various versions for our SMT systems. Each experiment involved a number of steps. The corpora were processed, including tokenization, cleaning, factorization, lowercasing, splitting, and final cleaning after splitting. Training data was processed, and the language model was developed. Tuning was performed for each experiment. Lastly, the experiments were carried out.

The baseline system testing was done using the Moses open source SMT toolkit with its Experiment Management System (EMS) (Wołk, Marasek, 2013b). The SRI Language Modeling Toolkit (SRILM) (Wołk, Marasek, 2013b) with an interpolated version of the Kneser-Key discounting (–interpolate –unk –kndiscount) was used for 5-gram language model training.

We used the MGIZA++ tool for word and phrase alignment. KenLM (Heafield et al., 2013) was used to binarize the language model, with a lexical reordering using the msd-bidirectional-fe model. The symmetrisation method was set to grow-diag-final-and for word alignment processing.

Starting from baseline systems (BASE) tests in the PL to EN and EN to PL directions, we improved translation score through:

- extending the language model (LM),

- interpolating it (ILM)

- extending corpora with additional data (EXT)

-filtering additional data with Modified Moore Levis Filtering (MML) (Koehn, Haddow, 2012).

It must be noted that the extension of language models was done on the systems with the corpora after MML filtration. The results of the experiments are showed in Table 4 and Table 5.

| Corpus | System | BLEU | NIST | TER | METEOR |
|---|---|---|---|---|---|
| TED | BASE | 16,96 | 5,26 | 67,10 | 49,42 |
| | EXT | 16,96 | 5,29 | 66,53 | 49,66 |
| | MML | 16,84 | 5,25 | 67,55 | 49,31 |
| | LM | 17,14 | 5,27 | 67,66 | 49,95 |
| | **ILM** | **17,64** | **5,48** | **64,35** | **51,19** |
| BTEC | BASE | 11,20 | 3,38 | 77,35 | 33,20 |
| | EXT | 12,96 | 3,72 | 74,58 | 38,69 |
| | MML | 12,80 | 3,71 | 76,12 | 38,40 |
| | LM | 13,23 | 3,78 | 75,68 | 39,16 |
| | **ILM** | **13,60** | **3,88** | **74,96** | **39,94** |
| EMEA | BASE | 62,60 | 10,19 | 36,06 | 77,48 |
| | EXT | 62,41 | 10,18 | 36,15 | 77,27 |
| | MML | 62,72 | 10,24 | 35,98 | 77,47 |
| | LM | 62,90 | 10,24 | 35,73 | 77,63 |
| | **ILM** | **62,93** | **10,27** | **35,48** | **77,87** |
| **EUP** | **BASE** | **36,73** | **8,38** | **47,10** | **70,94** |
| | EXT | 36,16 | 8,24 | 47,89 | 70,37 |

|  | MML | 36,66 | 8,32 | 47,25 | 70,65 |
|---|---|---|---|---|---|
|  | LM | 36,69 | 8,34 | 47,13 | 70,67 |
|  | ILM | 36,72 | 8,34 | 47,28 | 70,79 |
| OPEN | BASE | 64,54 | 9,61 | 32,38 | 77,29 |
|  | EXT | 65,49 | 9,73 | 32,49 | 77,27 |
|  | MML | 65,16 | 9,62 | 33,79 | 76,45 |
|  | LM | 65,53 | 9,70 | 32,94 | 77,00 |
|  | **ILM** | **65,87** | **9,74** | **32,89** | **77,08** |

Table 4. Polish to English MT experiments

| Corpus | System | BLEU | NIST | TER | METEOR |
|---|---|---|---|---|---|
| TED | BASE | 10,99 | 3,95 | 74,87 | 33,64 |
|  | EXT | 10,86 | 3,84 | 75,67 | 33,80 |
|  | MML | 11,01 | 3,97 | 74,12 | 33,77 |
|  | LM | 11,54 | 4,01 | 73,93 | 34,12 |
|  | **ILM** | **11,86** | **4,14** | **73,12** | **34,23** |
| BTEC | BASE | 8,66 | 2,73 | 85,27 | 27,22 |
|  | EXT | 8,46 | 2,71 | 84,45 | 27,14 |
|  | MML | 8,50 | 2,74 | 83,84 | 27,30 |
|  | LM | 8,76 | 2,78 | 82,30 | 27,39 |
|  | **ILM** | **9,13** | **2,86** | **82,65** | **28,29** |
| EMEA | BASE | 56,39 | 9,41 | 40,88 | 70,38 |
|  | EXT | 55,61 | 9,28 | 42,15 | 69,47 |
|  | MML | 55,52 | 9,26 | 42,18 | 69,23 |
|  | LM | 55,38 | 9,23 | 42,58 | 69,10 |
|  | ILM | 55,62 | 9,30 | 42,05 | 69,61 |
| **EUP** | **BASE** | **25,74** | **6,54** | **58,08** | **48,46** |
|  | EXT | 24,93 | 6,38 | 59,40 | 47,44 |
|  | MML | 24,88 | 6,38 | 59,34 | 47,40 |
|  | LM | 24,64 | 6,33 | 59,74 | 47,24 |
|  | ILM | 24,94 | 6,41 | 59,27 | 47,64 |
| **OPEN** | **BASE** | **31,55** | **5,46** | **62,24** | **47,47** |
|  | EXT | 31,49 | 5,46 | 62,06 | 47,26 |

|     |       |       |       |       |
| --- | ----- | ----- | ----- | ----- |
| MML | 31,33 | 5,46  | 62,13 | 47,31 |
| LM  | 31,22 | 5,46  | 62,61 | 47,29 |
| ILM | 31,39 | 5,46  | 62,43 | 47,33 |

Table 5. English to Polish MT experiments

The results shown in Table 4 and Table 5, specifically the BLEU, Meteor and TER values in the TED corpus, were checked for relevant differences. We measured the variance due to the BASE and MML set selection. It was calculated using bootstrap resampling5 for each test run. The result for BLEU was 0.5, and 0.3 and 0.6 for METEOR and TER respectively. The results over 0 mean that there is a significant difference between the test sets and it indicates that a difference of this magnitude is likely to be generated again by a random translation process, which would most likely lead to better translation results in general. (Clark, Dyer, Lavie, Smith, 2011)

In order to verify above conclusion, we decided to train an SMT system using only data extracted from comparable corpora (not using the original in domain data). The mined data were used also as a language model. The evaluation was conducted on the same test sets that were used in Tables 4 and 5. We wanted to check how such a system would cope with a translation of domain specific text samples. This experiment would possibly verify the influence of additional data on translation quality and analyze the similarity between mined data and in-domain data. Table 6 and 7 present these results. The rows named BASE show the results for baseline systems trained on original in-domain data, the rows named MONO show systems trained only on mined data in one direction, and finally the rows named BI present the results for system trained on data mined in two directions with duplicate segments removed.

| Corpus | System | BLEU  | NIST | TER   | METEOR |
| ------ | ------ | ----- | ---- | ----- | ------ |
| TED    | BASE   | 16,96 | 5,24 | 67,04 | 49,40  |
|        | MONO   | 10,66 | 4,13 | 74,63 | 41,02  |
|        | BI     | 11,90 | 4,13 | 74,59 | 42,46  |
| BTEC   | BASE   | 8,66  | 2,73 | 85,27 | 27,22  |
|        | MONO   | 8,46  | 2,71 | 84,45 | 27,14  |
|        | BI     | 8,50  | 2,74 | 83,84 | 27,30  |
| EMEA   | BASE   | 56,39 | 9,41 | 40,88 | 70,38  |

---

5 https://github.com/jhclark/multeval

|  | MONO | 13,72 | 3,95 | 89,58 | 39,23 |
|  | BI | 14,07 | 4,05 | 89,12 | 40,22 |
| EUP | BASE | 25,74 | 6,54 | 58,08 | 48,46 |
|  | MONO | 15,52 | 5,07 | 7155 | 51,01 |
|  | BI | 16,61 | 5,24 | 71,08 | 52,49 |
| OPEN | BASE | 31,55 | 5,46 | 62,24 | 47,47 |
|  | MONO | 9,90 | 3,08 | 84,02 | 32,88 |
|  | BI | 10,67 | 3,21 | 83,12 | 34,35 |

Table 6. PL to EN translation results using bi-directional mined data

| Corpus | System | BLEU | NIST | TER | METEOR |
|---|---|---|---|---|---|
| TED | BASE | 9,97 | 3,87 | 75,36 | 32,82 |
|  | MONO | 6,90 | 3,09 | 81,21 | 27,00 |
|  | BI | 7,14 | 3,18 | 78,83 | 27,76 |
| BTEC | BASE | 8,66 | 2,73 | 85,27 | 27,22 |
|  | MONO | 8,46 | 2,71 | 84,45 | 27,14 |
|  | BI | 8,76 | 2,78 | 82,30 | 27,39 |
| EMEA | BASE | 56,39 | 9,41 | 40,88 | 70,38 |
|  | MONO | 13,66 | 3,95 | 77,82 | 32,16 |
|  | BI | 13,64 | 3,93 | 77,47 | 32,83 |
| EUP | BASE | 25,74 | 6,54 | 58,08 | 48,46 |
|  | MONO | 9,92 | 4,10 | 72,51 | 32,06 |
|  | BI | 9,35 | 4,02 | 72,54 | 31,65 |
| OPEN | BASE | 31,55 | 5,46 | 62,24 | 47,47 |
|  | MONO | 6,32 | 2,23 | 92,40 | 22,72 |
|  | BI | 6,53 | 2,27 | 89,03 | 22,94 |

Table 7. EN to PL translation results using bi-directional mined data

The results of SMT systems based only on mined data were not surprising. Firstly, they confirm the quality and a high level of parallelism of the corpora that can be concluded from the high translation quality measured during experiments, especially for the TED data set. Only a two- BLEU-point gap can be observed when comparing the systems trained on the strict in-domain (TED) data and the mined data, when it comes to the EN – PL translation system. It also seems natural that the best SMT scores were obtained on the TED data. It is not only most similar to the Wikipedia articles and overlaps with it in many topics, but also the Yalign's classifier trained on the TED data set recognized most of parallel sentences. In consequence it can also be observed that the METEOR metric rises in some cases whereas other metrics decrease. The most likely reason for this is the fact that other metrics suffer, in comparison to the METEOR, from the lack of scoring mechanism for synonyms. Wikipedia is very rich not only when we consider its topics but also its vocabulary, which leads to a conclusion that mined corpora are a good source for extending sparse text domains. It is also the reason why the test sets originating from wide domains outscore narrow-domain ones. In addition, it is the most likely explanation why sometimes training on larger mined data slightly decreases results on test sets from very specific domains. Nonetheless, it must be noted that after a manual analysis we conceded that in many cases translations were good but automatic metric became lower because of the usage of synonyms. We also confirm once more that bi-directional mining has a positive influence on the output corpora.

Using the corpus of sentences generated with the analogy detection method, we obtained results presented in Table 8. We used the TED corpus for the experiments. Expanding the corpus with newly-generated sentences gave decreased results for all metrics. We seek a reason of this phenomenon and as a solution we tried to use sentences generated by the analogy method as a training corpus. The results of the experiment with the corpus obtained by this approach are presented in Table 8.

| PL-EN | BLEU | NIST | TER | MET |
|---|---|---|---|---|
| **TED Baseline** | 19,69 | 5,24 | 67,04 | 49,40 |
| **Analogy corpus** | 16,44 | 5,15 | 68,05 | 49,02 |
| **EN-PL** | **BLEU** | **NIST** | **TER** | **MET** |
| **TED Baseline** | 9,97 | 3,87 | 75,36 | 32,82 |
| **Analogy corpus** | 9,74 | 3,84 | 75,21 | 32,55 |

Table 8. Results on the TED corpus trained with an additional analogy based corpus

As a reason of such results we conclude that the analogy method is designed to extend existing parallel corpora from non-parallel data available. However, in order to establish a meaningful baseline, we decided to test a noisy-parallel corpus mined independently using this method. Therefore, the results are less favorable then the ones obtained using the Yalign method. Had we done otherwise, filtering effects would not have shown up in the test scores as the corpora differ significantly in size. As a solution to this problem, we decided to apply two different methods of filtering described in more detail in section 7.

## 7. Discussion and conclusions

Nowadays, bi-sentence extraction is becoming increasingly popular in unsupervised learning for numerous specific tasks. The method overcomes the disparities between English and Polish or any other West-Slavic languages. It is a language independent method that can easily be adjusted to a new environment, and it only requires parallel corpora for initial training. The experiments show that the method performs well. The obtained corpora increased the MT quality in wide text domains. A decrease or very small score differences in narrow domains are understandable because such a wide text domain as Wikipedia most likely adds unnecessary n-grams to a very specific domain that do not exist in test sets. Nonetheless, we can assume that even small differences can have a positive influence on real-life rare translation scenarios. In addition, we have demonstrated that mining data using two classifiers trained from a foreign to native (PL to EN) language and in the opposite direction (EN to PL) can significantly improve the quantity of the mined data even if some repetition occurs. Such bi-directional mining, which is logical, found additional data mostly for domains if wide range. In narrow text domains, the potential gain was not worth the effort. From the practical point of view, the method neither requires expensive training nor requires language-specific grammatical resources, while producing satisfactory results. We are able to replicate such mining for any language pair or text domain.

Nevertheless, there is still some room for improvement in two areas. In the presented experiments the amount of obtained data is not completely satisfactory. It must be mentioned that the classifier that was trained on the wide TED Talks corpora provided the biggest parallel corpus. When the classifier was trained with corpora from other narrow domains, like e.g. proceedings of the European parliament, medical texts, etc., the results of mining differed in size and content. The texts were narrowed just to the scope of one specific domain. Although a small improvement in translation quality was demonstrated, the limitation of the classifier domain provided data that did not extend the original corpora, as we had anticipated. Because of that it is of interest to train universal models and combine extracted corpora together in order to cover more translation scenarios. Moreover, developing a tuning script for acceptance parameters in the Yalign tool would most likely provide better results.

Unfortunately, it has to be noted that the final corpora contain noisy data. They contain mostly good translations but also some badly-aligned ones as well as some that are about the same topic but the translation is far too indirect to improve the MT quality. This is also the most likely reason for a small decrease in translation quality for tnarrow text domains. Filtering out such noisy data would certainly improve the influence of corpora on translations. We are currently working on a tool that should be able to filter such data.

Our strategy is to find a correct translation of each Polish line using any translation engine. We translate all lines of the Polish file (src.pl) with a translator and put each translation line in an intermediate English translation file (src.trans). This intermediate translation helps us find the correct line in the English translation file (src.en) and put it in the correct position or remove incorrect pairs from the corpora. However, there are additional complexities that must be addressed. Comparing the src.trans lines with the src.en lines is not easy, and it becomes harder when we want to use the similarity rate to choose the correct, real-world translations. There are many strategies to compare two sentences. We can split each sentence into tokens and find the number of words in both sentences. However, this approach has some problems. For example, let us compare "It is origami." to these sentences: "The common term origami is about how we use paper to create a form from it." and "This is origami.". Is such a case a sentence "This is origami." would be considered as less similar, which is obviously wrong.

Firstly, it is necessary to deal with stop words before comparing two sentences. Another problem is that sometimes we find words with the same stem in sentences, for example "boy" and "boys." The next comparison problem is word order in sentences, which is free in the Polish language. During the comparison phase synonyms should also be taken into account.

For finding equivalent words we used the NTLK Python module and WordNet[6] in order to find synonyms for each word and to use these synonyms in comparing sentences. Using synonyms for each word, we created multiple sentences from each original sentence and compared them as a many-to-many relation.

To obtain the best results, our script makes it possible to have multiple functions with multiple acceptance rates. Fast functions with lower quality results are tested first. If they can find results with a very high acceptance rate, we accept their selection. If the acceptance rate is not sufficient, we use slower but higher accuracy functions (Wołk, Marasek, 2014b).

The data is quite noisy and the corpora contain redundant parallel lines that contain just numbers or symbols. Additionally, it is easy to find improper translations e.g. "U.S. Dept." is surely not a translation of the sentence "Na początku lat 30", which in Polish means "At the beginning of the 30s". What is more, some translations are too indirect or too distinct from each other. An example to such a pair can be "In all other cases it is true." and "W przeciwnym razie alternatywa zdań jest fałszywa.", which in Polish means "Otherwise, the alternative of the sentences is false.".

---

[6] http://www.nltk.org/howto/wordnet.html

Although most of the corpora contain good translations, the problematic data should be removed. We conducted an initial experiment based on 1,000 randomly selected bi-sentences from the corpora. The data was processed by our filtering tool. Most of the noisy data was removed, but also some good translations were lost. Nevertheless, results are promising and we intend to filter the entire corpora in the future. It also must be noted that the filtering tool was not adjusted to this specific text domain. The results are presented in Table 9.

| | |
|---|---|
| Number of sentences in the base corpus | 1000 |
| Number of poor sentences in the test corpus | 182 |
| Number of filtered poor sentences | 154 |
| Number of filtered good sentences | 12 |

Table 9. Initial filtering results

We do not find the analogy-based results satisfactory. The reason is the low quality of the newly generated corpus. In our opinion the problem is that, in contrast with the Yalign method, the analogy-based method does not mine domain specific data. Additionally, we noticed that it suffers from duplicates and a relatively big amount of noisy data. As a solution to this problem, we decided to apply two different methods of filtering. The first one is easy, based on length of sentences in a corpus. We removed duplicates and very short (fewer than 10 characters) sentences as non-significant. As a result, we obtained 58,590 sentences in the corpus. We report the results in Table 11 as FL1 results. Secondly, we applied the filtration method described above (FL2). The results are showed in Table 11. The number of unique EN tokens before filtration was equal to 137,262 and PL to 139,408, after filtration we obtained 28,054 and 22,084 unique tokens respectively. Such filtrations improved SMT results concerning the analogy-based corpora showed in Table 11.

| | |
|---|---|
| **Number of sentences in the base corpus** | **3 800 000** |
| **Number of rewriting models** | 8128 |
| **Bi- sentences in the base corpus** | 114107 |
| **Bi-sentences after duplicates removal** | 64080 |
| **Remaining bi-sentences after filtration (FL1)** | 58590 |

| | **Remaining bi-sentences after filtration (FL2)** | 6557 |
|---|---|---|

Table 10. Filtration results of the analogy-based method (number of bi-sentences)

In order to evaluate the influence of filtration on the analogy-based corpora we trained SMT systems for each of the domains described above. The low SMT results confirmed our opinion that the obtained corpus is not domain specific and that it can be used for general purposes. The results are presented in Table 11. The row meanings are the same as in Table 10. An interesting fact is that the EMEA test set provided higher baseline and filtered results. The source of such a phenomenon can be attributed to the similarity between the textual content of the Wikipedia and EMEA corpora.

| | PL-EN | | | | EN-PL | | | |
|---|---|---|---|---|---|---|---|---|
| | **BLEU** | **NIST** | **TER** | **MET** | **BLEU** | **NIST** | **TER** | **MET** |
| **TED** | | | | | | | | |
| **Analogy corpus** | 1,87 | 1,55 | 93,92 | 17,88 | 0.91 | 0.97 | 99.68 | 10.77 |
| **FL1** | 1,26 | 1,02 | 87,94 | 14,15 | 0.96 | 1.02 | 99.48 | 11.19 |
| **FL2** | 1,91 | 1,70 | 91,62 | 18,98 | 1.02 | 0.97 | 94.45 | 11.40 |
| **EUP** | | | | | | | | |
| **Analogy corpus** | 3,35 | 1,96 | 94,49 | 22,63 | 2.06 | 1.38 | 96.44 | 12.88 |
| **FL1** | 2,08 | 1,49 | 90,21 | 13,67 | 2.08 | 1.49 | 90.21 | 13,67 |
| **FL2** | 2,64 | 1,79 | 90,53 | 20,08 | 1.90 | 1.24 | 99.21 | 12.82 |
| **EMEA** | | | | | | | | |
| **Analogy corpus** | 5,75 | 2,16 | 99,19 | 22,01 | 8.61 | 2.50 | 89.99 | 20.83 |
| **FL1** | 8,75 | 2,59 | 87,40 | 21,69 | 8.75 | 2.59 | 87.40 | 21,69 |
| **FL2** | 8,08 | 2,46 | 97,39 | 23,19 | 9.45 | 2.54 | 88.59 | 22.01 |
| **OPEN** | | | | | | | | |
| **Analogy corpus** | 1,41 | 1,12 | 104,60 | 14,06 | 2.40 | 0.92 | 98.03 | 11.17 |
| **FL1** | 1,20 | 0,93 | 98,58 | 11,77 | 1.20 | 0.93 | 98.58 | 11.77 |
| **FL2** | 3,15 | 1,28 | 98,30 | 11,77 | 2.47 | 1.03 | 97.31 | 12.9 |

Table 11. Results in SMT on analogy based sentences, filtrated corpus: FL1, FL2